\documentclass[sigconf]{acmart}
\AtBeginDocument{%
  \providecommand\BibTeX{{%
    \normalfont B\kern-0.5em{\scshape i\kern-0.25em b}\kern-0.8em\TeX}}}

\setcopyright{acmcopyright}
\copyrightyear{2018}
\acmYear{2018}
\acmDOI{10.1145/1122445.1122456}

\acmConference[Worldwide online '21]{Worldwide online '21: ACM Interaction Design and Children (IDC) conference}{June 24--30, 2021}{Worldwide online}
\acmBooktitle{Worldwide online '21: ACM Interaction Design and Children (IDC) conference workshop,
  June 24--30, 2021, Worldwide online}
\acmPrice{15.00}
\acmISBN{978-1-4503-XXXX-X/18/06}

\begin{document}

\title{Designing Games for Enabling Co-creation with Social Agents}

\author{Safinah Ali}
\authornote{Both authors contributed equally to this research.}
\email{safinah@media.mit.edu}
\author{Nisha Devasia}
\authornotemark[1]
\email{ndevasia@media.mit.edu}
\affiliation{%
  \institution{MIT Media Lab}
  \streetaddress{75 Amherst Street}
  \city{Cambridge}
  \state{Massachusetts}
  \country{USA}
  \postcode{02142}
}

\author{Cynthia Breazeal}
\affiliation{%
  \institution{MIT Media Lab}
  \streetaddress{75 Amherst Street}
  \city{Cambridge}
  \state{Massachusetts}
  \country{USA}
\email{cynthiab@media.mit.edu}
}

\renewcommand{\shortauthors}{Ali and Devasia, et al.}

\begin{abstract}
Digital tools have long been used for supporting children's creativity. Digital games that allow children to create artifacts and express themselves in a playful environment serve as efficient Creativity Support Tools (or CSTs). Creativity is also scaffolded by social interactions with others in their environment. In our work, we explore the use of game-based interactions with a social agent to scaffold children's creative expression as game players. We designed three collaborative games and play-tested with 146 5-10 year old children played with the social robot Jibo, which affords three different kinds of creativity: verbal creativity, figural creativity and divergent thinking during creative problem solving. In this paper, we reflect on game mechanic practices that we incorporated to design for stimulating creativity in children. These strategies may be valuable to game designers and HCI researchers designing games and social agents for supporting children's creativity.
\end{abstract}




\maketitle

\section{Introduction}
Digital tools such as phones, tablets, computers, and tangible devices have been used as \textit{creativity support tools} (CSTs) for over two decades ~\cite{frich2019mapping}. 
Video games have proven an effective medium for scaffolding creativity \cite{Wallach65, Koster2013-yb, Ott2012-il} due to a variety of game design choices, such as optimal challenges, risk-taking, exploration, and the ability to learn by failing \cite{Kim2015-ye}. Recent work has also started leveraging social agents such as social robots as creativity support tools, using interactive activities such as dance ~\cite{Ros2013-da}, storytelling ~\cite{Ligthart2020-ob, Alves-Oliveira2020-bw}, and playing music ~\cite{McCallum2015-wl} to engage children in creative expression. Children learn creativity from their peers in a collaborative space. However, these benefits may be lost while using digital learning tools designed for a single learner. As robots begin to enter personal living spaces, as well as being used increasingly as pedagogical tools~\cite{Belpaeme2018-zr}, there is a unique opportunity to use their social abilities, as well as the connections humans form with them, to promote collaboration, learning, and creative thinking. Social robots are uniquely situated in their ability to express socially and emotionally, while acting as a peer, to engage children in social interactions that can stimulate creative development. In our work, we explore how game-based interactions with social interactive agents can help support children's creativity. We designed three two-player games that children play collaboratively with the social robot Jibo. The Droodle creativity game affords verbal creativity where children and the robot take turns to come up with witty captions for abstract images. The MagicDraw game affords figural creativity where the child and the robot collaboratively complete drawings. The Escape!Bot game affords divergent thinking and creative problem solving where the children and the robot play a platform game that involves escaping using creative combinations of physical contraptions. 

In this paper, we reflect on our learnings by providing game design recommendations for child-robot co-play that we found effective in fostering creativity in children.

\section{Background}
Previous research in game design, human-computer interaction and human-robot interaction has outlined how games can be beneficial for creative growth, and how social interactions by social agents can foster creativity in children. 

\subsection{Games for Creativity}
Game-like activities and play-based learning tools have been shown to enhance creativity \cite{Wallach65}. Video games afford the facilitation of creative behavior and risk taking by providing optimal challenges and sandbox-like exploration to players \cite{Rahimi_undated-xs}. They are also effective vehicles for creative learning \cite{Koster2013-yb}, and bring elements of engagement and fun into the learning process, which can lead to creative problem solving \cite{Hutton2010-tv}. Providing the correct amount of challenge within a game-based task to a user aids in absorbing them in the process, leading to a state commonly referred to as flow ~\cite{Nakamura2014-jj}. Henriksen discusses \textit{assimilation} (the process of adding facts to existing mental structures or schemas), \textit{accommodation} (the process of changing one’s personal interpretative frame when it has become perturbated with
knowledge that renders it invalid), and \textit{perturbation} (allowing the participant to experiment with an alternative and temporary way of experiencing the world) as methods of constructing creative learning environments within games \cite{Henriksen_TD2006-rf}. Ott and Pozzi demonstrated that playing digital games appreciably increased students' creative attitudes and skills in the long term \cite{Ott2012-il}. 

In our studies, we designed interactive digital games for Android tablets. Children play these games together with the social robot, and create artifacts as well as creatively problem solve. 

\subsection{Social Agents and Creativity} 
Children learn from other creators in their environments, such as teachers and classmates, through mechanisms of social emulation \cite{Whiten2009-lb}. Social interactions with teachers and peers have been shown to positively influence children's creativity \cite{Kafai1995-rg}. However, little work has been done on leveraging the social abilities of social agents to scaffold and enhance children's creativity. In this work, we utilize social robots, which have proven to be effective tools for pedagogical and cognitive gains \cite{Belpaeme2018-zr}. Kahn et. al demonstrated how interaction with a social robot can help generate more creative ideas \cite{Kahn2016-ma}. Social robots can also be effective peers that lead to measurable learning gains \cite{Leyzburg14}, and can promote positive learning behaviors in children through social emulation, such as curiosity, verbal creativity, and growth mindset \cite{Gordon15, Ali19, Park17}. Our works utilize Jibo, an tabletop social robot that can speak, display expressions through its screen and body movements, respond to speech, and track sound and movement. Jibo takes the place of a co-present, collaborative peer during gameplay.

In our work, we combine these two approaches are explore how social interactions with social agents during collaborative gameplay can stimulate creativity. We designed three game based child-robot interactions where children play the games with the robot acting as a collaborative peer. The games afford different kinds of creativity where the gameplay itself promotes creative expression and the robot scaffolds children's creativity by itself demonstrating creative behaviors or social interactions such as asking questions, suggesting ideas and providing positive reinforcement.

\section{Games Designed for Creative Child-Robot Interactions}

In this section we describe the gameplay of the three child-robot collaborative games.

\subsection{Droodle Creativity Game}
In order to afford verbal creativity, we designed the Droodle Creativity game \cite{Ali19}, inspired by the Droodle Creativity Task~\cite{kahn2005creativity}, a verbal creativity task that draws upon people’s ability to find creative ways to describe an abstract image or figure known as a droodle. In the game, two players take turns generating Droodle titles. The active player is presented with droodles on a tablet screen and they come up with droodle title(s) in 30 seconds. The goal is to come up with witty titles for abstract droodles. Players then switch turns until each player has played five turns each, which takes about 20 minutes in total. 

To stimulate creativity in the children, the robot itself demonstrated creative behaviors. Literature in HRI demonstrates that children socially emulate social robots' behaviors~\cite{Park17, zanatto2020humans}. To demonstrate high creativity, the robot picks many titles (fluency), unique topics (novelty) and highly \textit{creative} titles (as determined by the Droodle creativity metric~\cite{kahn2005creativity}) from a corpus of Droodle titles. Further, the game does not involve a fail state or scoring metric. Players keep generating titles in the time allotted and the next player take the turn. The robot supports the child in their turn by providing affirmation every time the child submits a title. For instance, after the child's response, the robot says, "Oh that is a good one. What else can it be?"

\begin{figure}[h]
\centering
\includegraphics[width=0.8\linewidth]{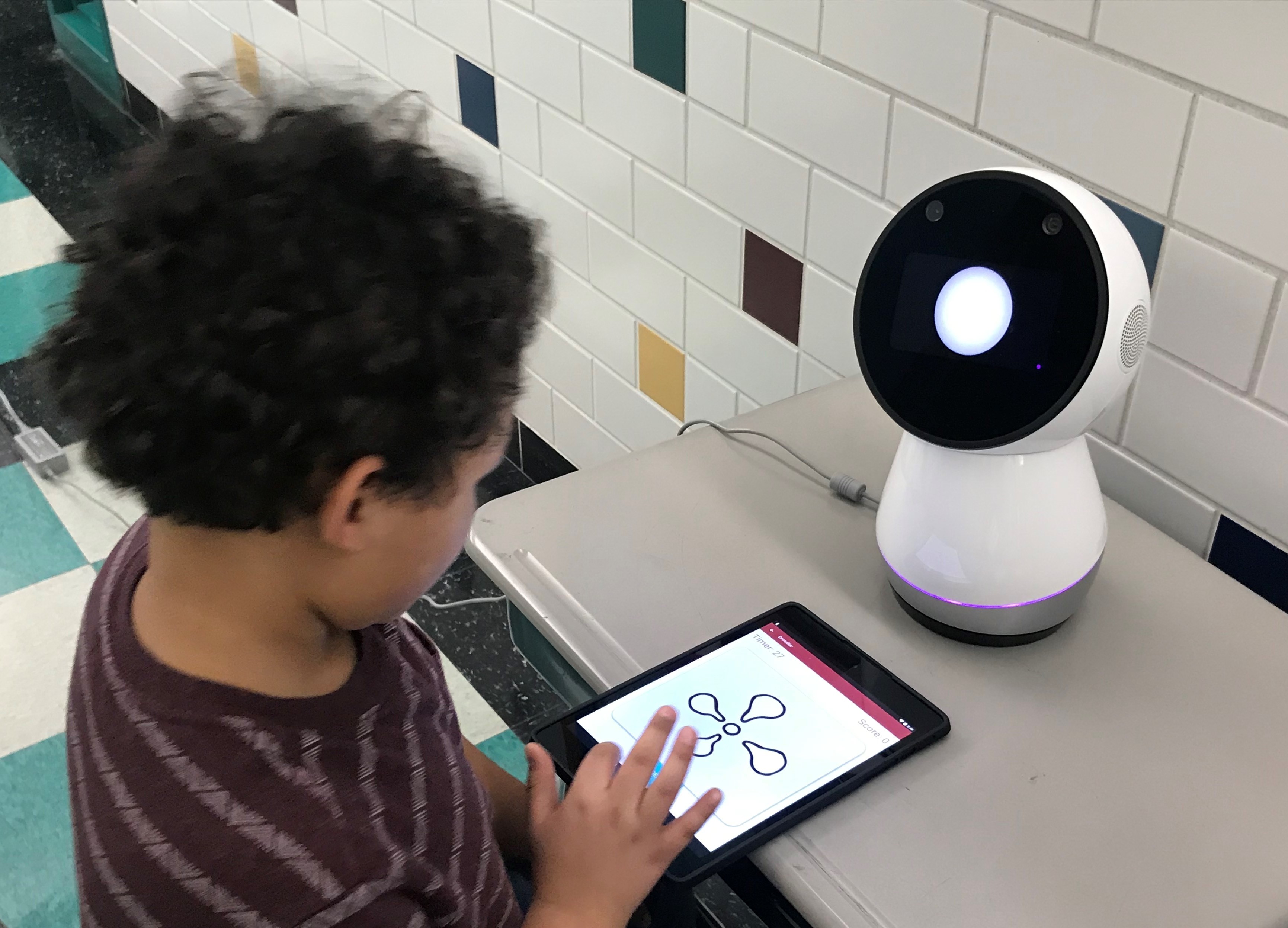}
\caption{A child looking at a Droodle in the Droodle Creativity Game.}
\label{fig:interface}
\end{figure}

\subsection{MagicDraw}
To afford children's figural creativity, we designed the MagicDraw game, which involves a collaborative drawing interaction on an Android tablet between the child and the robot \cite{Ali2020-rc}. The gameplay requires one player to start a drawing with one stroke, and the other player will transform the drawing into a meaningful object. After the drawing is complete, the players switch turns. When it is the robot’s turn to draw, we utilize the Sketch-RNN drawing model ~\cite{Ha2017-om}, which converts a starting stroke into a meaningful illustration chosen from a set of categories. Playing through all the turns takes about 20 minutes. 

We chose the drawing interaction not only to afford figural creativity but also because there are several ways to represent an object and it allows for creative freedom to explore many drawings. The game interaction lets each player draw multiple times. In the robot's turn, the robot makes multiple drawing attempts and keep improving the drawing quality. The robot also reflects on the drawing strategy. For instance, when the child made two lines as a starting prompt and indicated \textit{flamingo} as the target category, the robot said, "I can use those as flamingo legs and make a body." Like the Droodle game, the MagicDraw game does not involve any assessment or score and the robot does not judge any drawing as incorrect. The robot provides positive reinforcement every time the child submits a drawing. Finally, the robot itself demonstrates figural creativity by converting the child's drawings into meaningful objects. The robot constantly asks for the child's input and feedback through questions like, "What do you think?", which make the experience collaborative and interactive for the child player. 

\begin{figure}[h]
\centering
\includegraphics[width=0.8\linewidth]{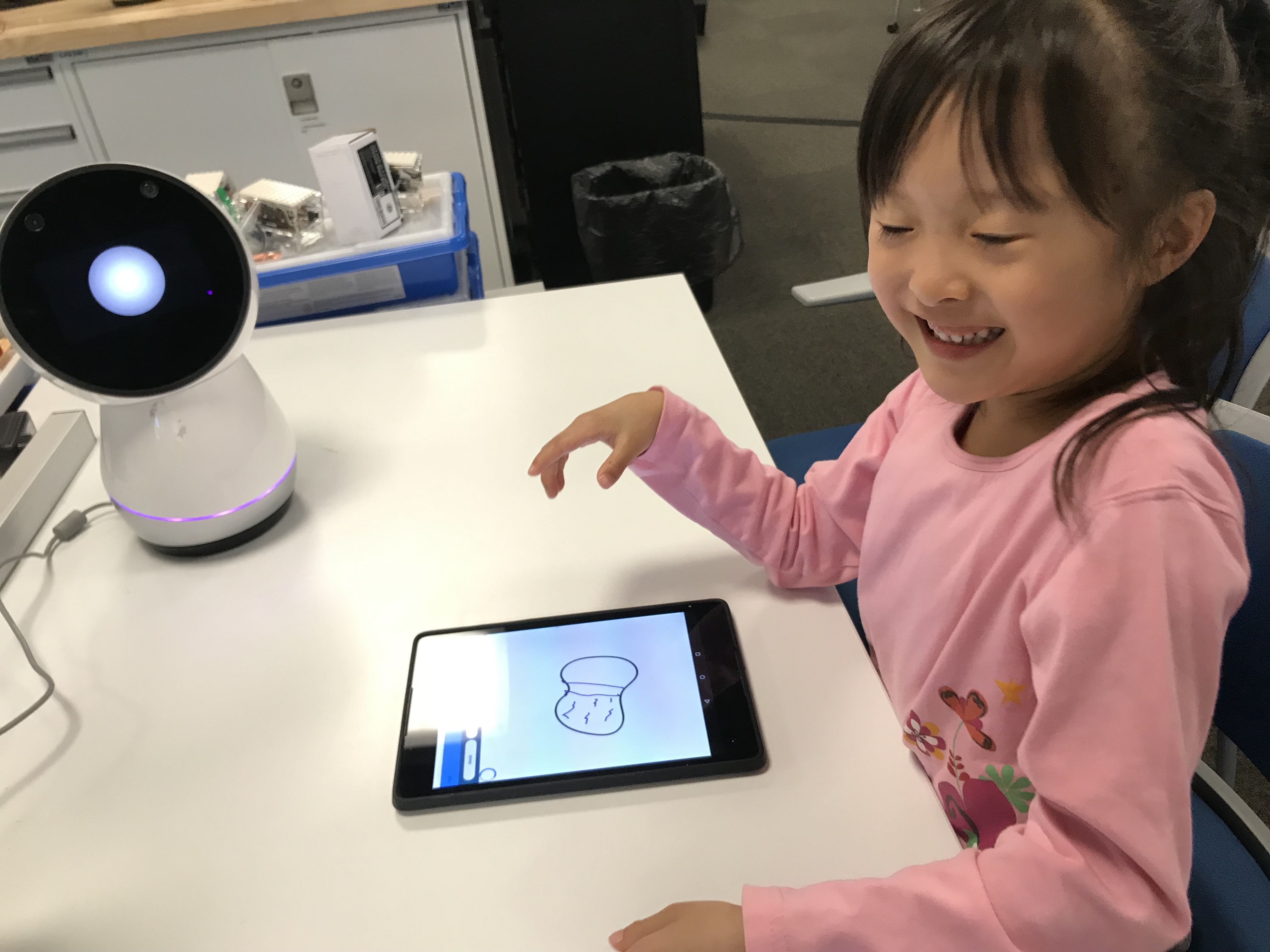}
\caption{A child drawing a butterfly in the MagicDraw game.}
\label{fig:interface}
\end{figure}
\subsection{Escape!Bot}
 Escape!Bot is a sandbox style tablet game that allows players to form strategies to escape the level by leveraging creative combinations of physical contraptions. The two players (child and robot) play the game collaboratively, and the robot provides the child with ideas, positive feedback and asks questions. Players must get the character to an end marker by placing combinations of tools that act as physical contraptions within the game space. The tools can be combined in any logical manner to allow for an unlimited number of ways to traverse through the level. There are three levels, and participants are given 45 minutes.

\begin{figure}[h]
\centering
\includegraphics[width=0.8\linewidth]{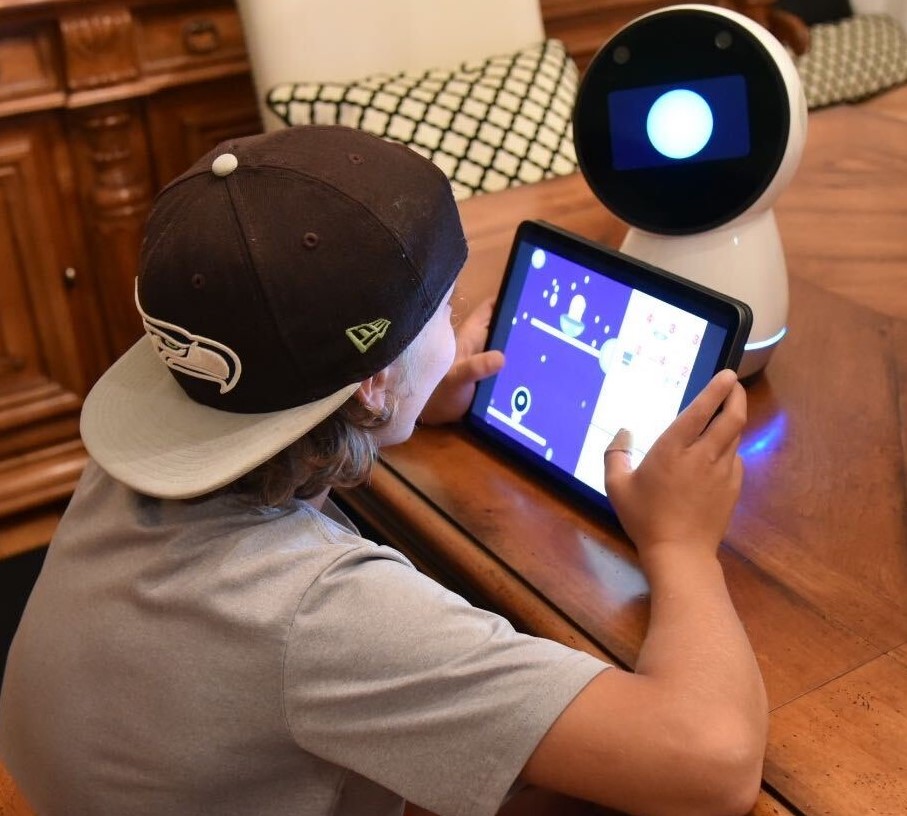}
\caption{An example of a child playing the Escape!Bot game in the embodied condition.}
\label{fig:interface}
\end{figure}

The robot's behaviors in Escape!Bot were inspired by Kahn et. al's 10 human-robot interaction patterns aimed to foster human creativity, and consist of three categories: demonstration (\textit{Pushing the Limits}), encouragement (\textit{Validate Decision}), and inquiry (\textit{Consider the Alternative}) \cite{Kahn2016-ma}. The robot \textit{demonstrates} creativity by providing the player with combinations of objects that they may not have previously considered. For instance, if the player is using a disproportionate number of ladders, the robot will say "I see you've been using a lot of ladder. Why don't you try combining it with other objects? I bet you could put ladder on top of a ramp! Or on top of a box!". Similarly, the robot \textit{encourages} creativity by positively reinforcing the use of varied objects as well as reusing objects. For example, the robot will praise the player if they use every different object by saying, "Knowing how all the objects work is useful!", and if the player reuses objects (a mechanic not specifically explained in the tutorial), the robot will say, "Reusing objects is smart!". Finally, the robot \textit{inquires} how the player could use consider using another object to achieve the same functionality as the one they are currently using. For instance, if the player runs out of ladders, the robot will randomly select an object that the player hasn't used with similar functionality and prompt them to use it.

\section{Discussion}
We playtested the games with 146 students (48 Droodle Creativity Game, 78 MagicDraw Game, 20 EscapeBot Game) in order to analyze the effect of the robots' social interactions on children's creativity. We found that social interactions expressed by the social robot such as creativity demonstration and creativity scaffolding led to creativity growth in children during the collaborative tasks~\cite{Ali19, Ali2020-rc, Ali2021-tu, Devasia2020-nj}. While detailed study findings of creativity gains are beyond the scope of this paper, in the next section, we discuss game design strategies in the three child-robot games that we incorporate and found to be successful. 

\subsection{Affording Creation of Artifacts}
The Droodle Creativity Game and MagicDraw Game were designed around the creation of artifacts, rather than the completion of a specific deliverable. Creativity literature shows that evaluation has a negative effect on creativity, and as such, the artifacts in these games had no correct or expected answers, which encouraged free and unlimited creative expression. For example, in the Droodle Creativity Game, participants could come up with any number of innovative titles for an image, and in the MagicDraw Game, participants create a doodle of their own devising. 

\subsection{Social Agents Demonstrating Creativity}
We observed that children can learn verbal and figural creativity by emulating a social robot’s creative behaviors. Children paired with the creative robot in the Droodle Creativity Game improvised a significantly larger number of valid Droodle titles than children who interacted with the non-creative counterpart. Similarly, children emulated the robot's creative drawings in the MagicDraw game, expressing higher levels of creativity than children in the non-creative condition. Therefore, in order to utilize social robots as effective pedagogical tools, it is essential that they express the desired creative behavior that researchers aim to foster in children. Creativity can also be supported by social behaviors such as reflecting on the artifact or the creative process itself.

\subsection{Social Agents as Collaborative Peer Players}
In all three of the games, we framed the social robot agent as a collaborative peer assisting the child in their task. Collaboration is a key social factor involved in positively influencing creativity, and as such, the agent's interactions must be carefully designed to ensure that children see it as a partner rather than a competitor. Dialogue from the robot introducing itself as well as the game can be leveraged to position the robot as a collaborative peer. For example, in the Escape!Bot game, participants are given a tutorial scene in which Jibo introduces itself and positions the participant as a partner trying to help it find its way home.

\subsection{Avoiding Assessment}
We deliberately chose to not include a scoring mechanic, nor any sort of feedback system, with regards to the player's solutions to the task, as assessment is shown to hinder creativity. Rather than judging or providing feedback, the robot is instead designed to express curious non-verbal behaviors, or verbal positive reinforcement, in response to the player's actions within the task. For example, in the Escape!Bot game, despite the fact that the game can be won, the robot does not offer any suggestions for optimal solutions, allowing the player to experiment with several different types of strategies.

\subsection{Supporting Creative Exploration}
All three games contain an endless possibility for the solutions that could be devised for the task. In the Droodle Creativity Game, participants were not provided with any constraints for possible answers, and similarly, the MagicDraw game allowed children to improvise a drawing from basic shapes. Similarly, the Escape!Bot game provides an unlimited number of permutations for reaching the end of the level. All games allow players multiple attempts to complete the task, and are only limited by the time constraints of the study session itself. This provides players with a sandbox like environment for exploration and experimentation, a modality that has been shown to enhance creativity. Further, the robot embodied this behavior by making multiple attempts at generative Droodle ideas or making multiple drawings in the MagicDraw game.

\subsection{Optimal Challenges}
Balancing the correct amount of difficulty is essential to entering the state of flow, in which players of a game are at their highest performance. Therefore, social agents must present children with optimal challenges and challenge their assumptions to positively affect their creativity. For example, in the Escape!Bot game, the robot would comment if a child was using a large amount of a particular object, and asked them to try using another object with similar functionality. These questions enabled students to explore creative uses of the objects beyond the first use that they imagined, which encouraged flexibility. 

\section{Limitations and Conclusions}
A limitation of the current work is the wide age range used, and future analysis would be needed to determine the efficacy of the intervention on different age groups. Another limitation of this work is that these tasks are narrow constructed, and do not fully represent the scope of human creativity. Creativity encompasses a much wider array of behaviors (such as divergent thinking) that can be explored using other interactions. Furthermore, while all these interactions currently focus on one-on-one child-robot interaction, we must strive towards multi-party interactions because collaboration with peers forms a major part of creative learning. This work also does not measure the efficacy of using social robots as creativity support tools in comparison with other digital modalities or even in comparison with interacting with human peers. Future work is required to understand the differences between human-human and human-robot creative collaboration. 

In this workshop, we hope to discuss game design techniques to promote learning behaviors in children such as creativity, as well as how social interaction with artificial agents can foster positive learning behaviors. In our work, we leveraged background work on the design of transformative games and creativity supporting games, and social agents to design game interactions. Further, we modified game mechanics and social interactions with agents after play-testing the game with students. From this workshop, we, we aim to learn co-design methodologies to not only design the games for children but \textit{with} children to  support their creativity.




\bibliographystyle{ACM-Reference-Format}
\bibliography{sample-base}


\end{document}